\pdfoutput=1
\documentclass[11pt]{article}

\usepackage[final]{acl}

\usepackage{times}
\usepackage{latexsym}

\usepackage[T1]{fontenc}
\usepackage[utf8]{inputenc}

\usepackage{microtype}

\usepackage{inconsolata}

\usepackage{graphicx}

\usepackage{xspace}
\usepackage{amsmath}
\usepackage{adjustbox,booktabs,multirow}

\newcommand{\methodname}{MEMIT-Merge\xspace}
\newcommand{\eg}{\emph{e.g.}\xspace}

\definecolor{mycolor}{RGB}{243,237,247}

\title{MEMIT-Merge: Addressing MEMIT's Key-Value Conflicts in Same-Subject Batch Editing for LLMs}

\author{Zilu Dong\thanks{Equal Contribution.},  Xiangqing Shen{\footnotemark[1]}, Rui Xia\thanks{Corresponding Author.}\\
        School of Computer Science and Engineering, \\ 
        Nanjing University of Science and Technology, China \\
        \{zldong, xiangqing.shen, rxia\}@njust.edu.cn}
\begin{document}
\maketitle

\begin{abstract}
As large language models continue to scale up, knowledge editing techniques that modify models' internal knowledge without full retraining have gained significant attention. 
MEMIT, a prominent batch editing algorithm, stands out for its capability to perform mass knowledge modifications.
However, we uncover that MEMIT's editing efficacy significantly deteriorates when processing batches containing multiple edits sharing the same subject. Our analysis reveals this stems from MEMIT's key value modeling framework: identical keys (derived from the shared subject) are forced to represent different values (corresponding to different knowledge), resulting in update conflicts during editing.
Addressing this issue, we propose MEMIT-Merge, an enhanced approach that merges value computation processes for facts sharing the same subject, effectively resolving the performance degradation in same-subject batch editing scenarios. 
Addressing this issue, we propose MEMIT-Merge, an enhanced approach that merges value computation processes for facts sharing the same subject, effectively resolving the performance degradation in same-subject batch editing scenarios. 
Experimental results demonstrate that when MEMIT's edit success rate drops to around 50\% at larger batch sizes, MEMIT-Merge maintains a success rate exceeding 90\%, showcasing remarkable robustness to subject entity collisions. The code is available at \url{https://github.com/NUSTM/MEMIT-Merge}.

\end{abstract}
\section{Introduction}
\label{sec:introduction}

% Knowledge editing techniques aim to update models' internal knowledge without retraining.
As large language models (LLMs) continue to scale up, the prohibitive cost of full model retraining has made knowledge editing increasingly crucial in this domain. 
% operates on the fundamental assumption that specific knowledge representations are localized within particular regions of the model's parameter space, enabling targeted modifications through precise manipulation of these identified regions.
Among prevalent editing algorithms, a class of algorithms, termed ``Locate and Edit'' methods by \citet{zhang_comprehensive_2024}, enables targeted modifications through precise manipulation of specific regions.
% MEMIT \cite{DBLP:conf/iclr/MengSABB23}, one of the most prominent algorithms in this class, has gained significant attention \cite{DBLP:conf/aaai/Li0SYMY24, fang_alphaedit_2024, gupta_unified_2024}.
% MEMIT inherits the core architectural feature of ROME \cite{DBLP:conf/nips/MengBAB22}, which localizes knowledge to specific layers and modifies the output linear layer of MLP modules to update knowledge. 
% The distinctive advancement of MEMIT lies in its capability to perform batch-wise mass knowledge editing, enabling simultaneous modification of multiple knowledge instances within a single batch.
MEMIT \cite{DBLP:conf/iclr/MengSABB23}, one of the most prominent algorithms in this class, has gained significant attention \cite{DBLP:conf/aaai/Li0SYMY24, fang_alphaedit_2024, DBLP:conf/emnlp/GuptaSA24}. 
%It extends ROME's architecture \cite{DBLP:conf/nips/MengBAB22} with mass-editing ability.
% While preserving ROME's core mechanism of localizing knowledge in MLP layers and modifying their output linear layers,
It extends ROME's architecture \cite{DBLP:conf/nips/MengBAB22} and enables the simultaneous modification of multiple knowledge instances within a single update operation.

\begin{figure}[t]
    \centering
    % \colorbox{mycolor}{
    \includegraphics[width=\columnwidth]{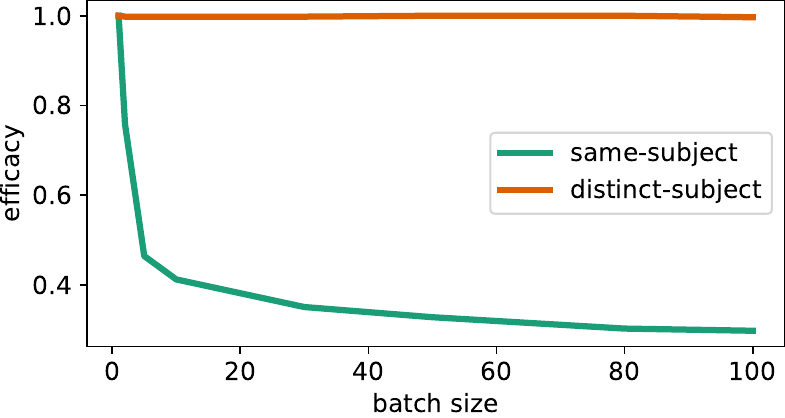}
                   \caption{The edit success rate of the MEMIT method on same-subject and distinct-subject datasets, showing the changes with varying batch sizes. A significant decline is observed when the subjects are the same. }
    \label{fig:same_diff_editsuccess}
\end{figure}

However, our investigation reveals a critical limitation in MEMIT: When handling batches with multiple edits that share the same subject (such as ``John Smith now plays basketball.'' and ``John Smith comes from England.'' share the same subject ``John Smith'', while ``Paul Morand comes from England'' has a different subject), the method will exhibit significant performance degradation. In contrast, edits with different subjects maintain stable efficacy.
% (such as ``Jack Johnson now plays basketball'' and ``Paul Morand comes from England'' with subjects `Jack Johnson'' and ``Paul Morand'', respectively) 

To systematically demonstrate this performance degradation, we constructed two contrastive datasets comprising batches with identical subjects versus fully unique subjects, named distinct-subject and same-subject, respectively. The experimental results are in Fig.~\ref{fig:same_diff_editsuccess}, where the vertical axis represents efficacy (which means the editing success rate) and the horizontal axis indicates the batch size per edit. The results reveal that MEMIT maintains a high success rate as batch size increases when editing distinct subject cases, but exhibits significant performance degradation for the same subject cases. However, same subject cases are also critical in the real-world practices (\eg, updating a person's occupation, workplace, and employer simultaneously).
More detailed experimental settings can be found in Sec.~\ref{subsec:same_diff_edit}.

The performance degradation stems from MEMIT's key-value modeling paradigm: identical keys (derived from shared subject representations) map to conflicting values during same-subject batch edits. 
MEMIT formulates knowledge updates as MLP key-value pairs where the output linear layer's weights are adjusted to align keys with edited values.
\footnote{Note that the key-value here refers to the hidden state and output within the MLP module as described by \citet{DBLP:conf/nips/MengBAB22}, rather than the query, key and value in the attention module.} 
However, when multiple edits share subjects, their identical keys require divergent value mappings - an inherent contradiction since single-layer perceptrons cannot produce multiple outputs for identical inputs. 
% This conflict escalates with increasing key similarity, directly correlating to performance deterioration.

To resolve this fundamental conflict, we propose \methodname, an enhanced variant of MEMIT. Our key insight is to enforce value consistency by merging multiple knowledge entries that share identical keys.
Experimental results show that \methodname consistently outperforms MEMIT on same-subject dataset, maintaining a success rate above 90\%, whereas MEMIT drops to around 50\%. For distinct-subject data, both methods perform comparably with no significant differences.

\section{Problem}

\subsection{Preliminaries}
\label{preliminaries}

The MEMIT framework hypothesizes that factual knowledge in models is stored within the parameters of MLP layers. Each MLP layer contains input/output linear layers with parameter matrices $W_{in}$ and $W_{out}$, where $W_{out}$ serves as the key-value mapping targeted by MEMIT editing. The key corresponds to the hidden state at the MLP’s intermediate layer  while the value represents the MLP’s final output.

Knowledge is represented as triples $(s, r, o)$. During editing, complete sentences are constructed from these triples. The key is determined by the subject $s$ and its contextual prefix, while the value is obtained by inversely optimizing the object $o$:
\begin{equation}
% \small
\label{eq:MEMIT_v_update}
v=\arg\min_v (-\log P_v [o|(s, r)])
\end{equation}
All $(k, v)$ pairs are processed in batch to update $W_{out}$ via closed-form solution:
\begin{equation}
W_{out} = W_0 +(V-W_0K)K^T(C+KK^T)^{-1}
\end{equation}
Here, $K$ and $V$ denote batched key/value matrices, $W_0$ represents original parameters, and $C$ is a knowledge-preservation constant.

\subsection{Same subject issue in MEMIT}
\label{sec:issues in MEMIT edits}

\begin{figure*}[htbp]
    \includegraphics[width=\linewidth]{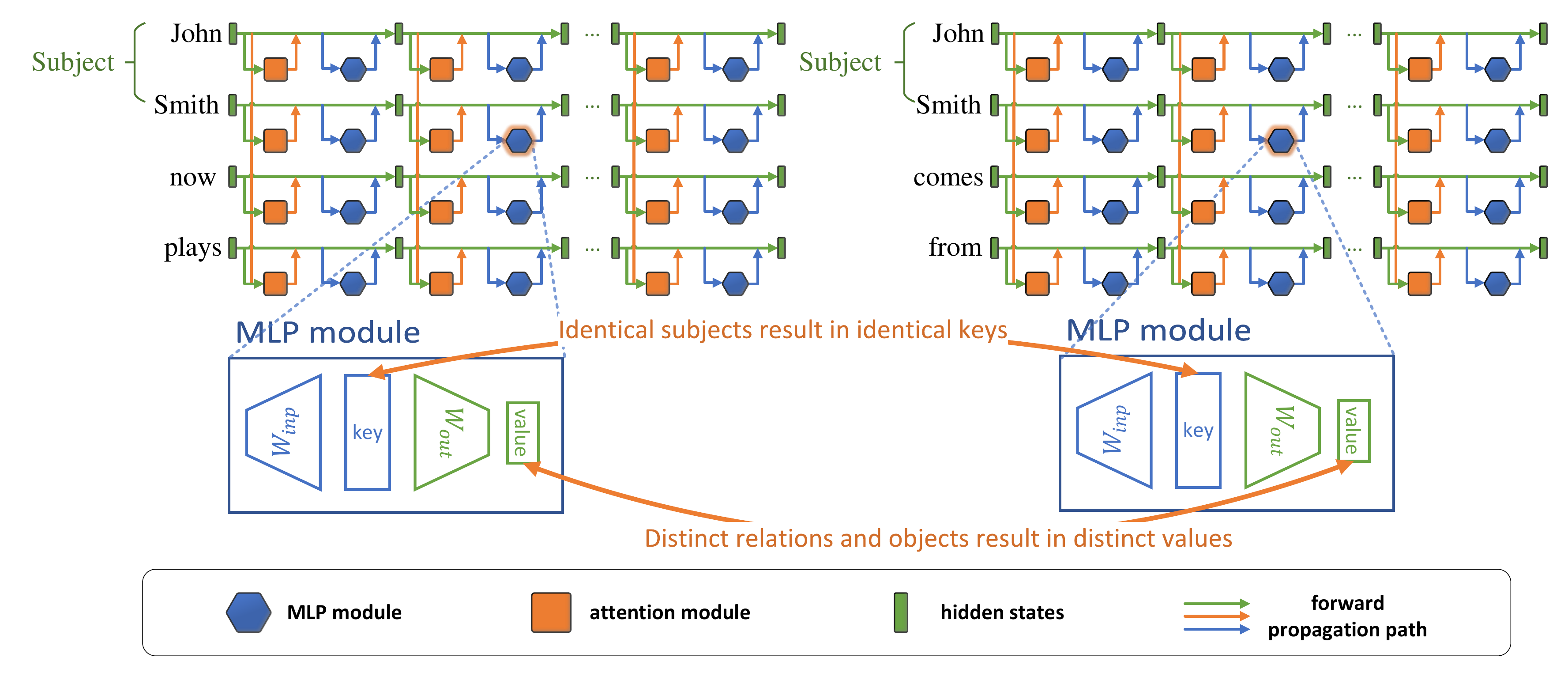}
    \caption{The architecture of MEMIT processing two same subject sentences. The left and right sides of the figure depict the processing flow of the two sentences respectively. Below, we expand the details of the MLP module to be modified, which consists of two linear layers. In MEMIT, the key is determined by the subject, resulting in identical keys on both sides. The value is optimized from the relation and object, leading to different values on each side. Consequently, the optimization target for the editable $W_{out}$ requires producing different values for the same input key.}
    \label{fig:general_process}
\end{figure*}

Normally, MEMIT is capable of maintaining its efficacy without a pronounced decline in performance when the edit batch size approaches 1,000.
% Moreover, the edit success rate remains relatively stable even when editing up to 10,000 knowledge triples.
However, we have identified a notable issue: when the edit batch encompasses knowledge triples sharing the same subject, the editing capacity of MEMIT experiences a substantial degradation.

To verify this phenomenon, we constructed two counterfactual editing datasets. In the first dataset, the subjects of the knowledge triples are all distinct. In the second dataset, the subjects of the knowledge triples are replaced by a single, fixed subject, while all other parts of these two datasets remain identical. The details of the construction of datasets are provided in App.~\ref{sec:appendix_dataset_construct}.

As illustrated in Fig.~\ref{fig:same_diff_editsuccess}, when the subjects are identical, the performance of the MEMIT method drops sharply with a batch size of only 2, and the edit success rate falls below 50\% when the batch size reaches 10. In contrast, when subjects are distinct, increasing the batch size has virtually no impact on edit success.

\section{Approach}
\label{sec:methods}

\subsection{Cause Analysis}
\label{sec:reason_analysis}

In our analysis, the degradation of editing capability caused by identical subjects is closely related to the key-value modeling of knowledge inherent in locate-and-edit class editing methods.

In the standard MEMIT, a piece of knowledge to be edited can be represented by a knowledge triple (subject, relation, object), and a complete sentence is constructed based on this triplet for the editing process. In this paper, we use the format ``{subject}'s {relation} is {object}'' to construct the sentence. For example, the knowledge triple (John,father,Bob) is formulated into the sentence ``John's father is Bob.''

As described in Sec.~\ref{preliminaries}, during MEMIT editing, the key is derived from the subject, while the value is determined by the object.
However, when editing multiple pieces of knowledge with the same subject but different objects in one batch, this mechanism forces the MLP to map the same key to two distinct values. As illustrated in Fig.~\ref{fig:general_process}, a given key can only produce a single fixed value through deterministic $W_{out}$. This creates a conflict when optimizing the parameter matrix, making it extremely challenging. We refer to this issue as the key collision problem. 
Consequently, when a batch contains multiple edits with the same subject, as demonstrated in Fig.~\ref{fig:same_diff_editsuccess}, the editing capability of MEMIT is significantly degraded.

Furthermore, we analyzed the relationship between MEMIT editing capability and key distance within a batch, finding that closer keys lead to greater capability degradation. Due to space constraints, detailed analysis is in the App.~\ref{sec:appendix_akd_analysis}

\subsection{The MEMIT-Merge Approach}
\label{sec:merge-approach}

To address this issue, we develop a new optimization objective to merge the value computation of the set of knowledge with the same key:

\begin{equation}
v=\arg\min_v \sum_{(s, r_j, o_j)\in S} -\log P_v[o_j|(s, r_j)], 
\end{equation}

where $S$ represents the set of knowledge triples with the same key,  $v$ is the value to be optimized in a backward manner, and $P_v$ denotes the model when the value is equal to $v$.  \footnote{For readability, the random prefix component, though part of the original MEMIT, is omitted from this equation. Nevertheless, it is incorporated identically to MEMIT in the real implementation.}

Compared with Eq.~\ref{eq:MEMIT_v_update}, this approach ensures that knowledge sharing same key gets the same value. 
As analyzed previously, original MEMIT encounters a contradiction: it demands that identical keys yield different values via the same $W_{out}$ matrix. Our method resolves this by merging.
Therefore, our approach can significantly alleviate the decrease in edit efficacy observed in standard MEMIT as evidenced in the next section.

Moreover, our methodology demonstrates superior theoretical efficiency compared to the original MEMIT. This advantage stems from the fact that our merging strategy effectively increases the batch size during gradient descent. In contrast to MEMIT's consistent batch size of 1 per edit, MEMIT-Merge dynamically sets its batch size to the number of edits sharing the same key.

\section{Experiments}

\begin{table*}[th]
\centering
\small
\begin{tabular}{@{}llllll@{}}
\toprule
Model                                   & Dataset                           & Method      & Efficacy & Paraphrase & Specificity \\ \midrule
\multirow{12}{*}{Qwen2.5-1.5B-Instruct} & \multirow{6}{*}{distinct-subject} & FT          & 0.24     & 0.21       & 0.99        \\
                                        &                                   & MEMIT       & 1.00     & 0.80       & 0.98        \\
                                        &                                   & PMET        & 0.94     & 0.67       & 0.98        \\
                                        &                                   & AlphaEdit   & 1.00     & 0.79       & 0.98        \\
                                        &                                   & PMET-Merge  & 0.94     & 0.67       & 0.97        \\
                                        &                                   & MEMIT-Merge & 1.00     & 0.80       & 0.98        \\ \cmidrule(l){2-6} 
                                        & \multirow{6}{*}{same-subject}     & FT          & 0.24     & 0.22       & 0.99        \\
                                        &                                   & MEMIT       & 0.41     & 0.28       & 1.00        \\
                                        &                                   & PMET        & 0.40     & 0.26       & 0.99        \\
                                        &                                   & AlphaEdit   & 0.42     & 0.27       & 0.98        \\
                                        &                                   & PMET-Merge  & 0.63     & 0.44       & 0.99        \\
                                        &                                   & MEMIT-Merge & 0.95     & 0.48       & 1.00        \\ \bottomrule
\end{tabular}
\caption{The complete results of the four editing methods—MEMIT, \methodname, PMET, AlphaEdit, PMET-Merge, and FT-L—on the same-subject and distinct-subject datasets at a batch size of 10. All experimental results were obtained by re-running each editing method on our dataset.}
\label{table:batch100_result}
\end{table*}

\subsection{Dataset}
\label{subsec:dataset}

We constructed two Wikidata-based counterfactual knowledge editing datasets: (1) a "same-subject" set with 100 triples sharing the subject John Smith, and (2) a "distinct-subject" set with unique subjects while maintaining identical relations/objects (construction details in App.~\ref{sec:appendix_dataset_construct}).

In terms of evaluation metrics, we refer to the metrics used by \citet{DBLP:conf/iclr/MengSABB23}, namely Efficacy, Paraphrase, and Specificity. Efficacy measures the edit success rate on original sentences, paraphrase measures the success rate on paraphrased sentences. Specificity measures the probability that facts unrelated to the edit remain consistent before and after the edit.

While our datasets are novel, they address critical real-world needs. Editing multiple attributes of an entity (\eg, updating a person’s profile) is a highly realistic demand, making same-subject scenarios essential for practical applications.

\subsection{Experimental Setup}
\label{subsec:setup}

We conducted experiments on three models with different architectures: Qwen2.5-1.5B-Instruct \cite{qwen_qwen25_2025}, GPT-J-6B \cite{gpt-j}, and Llama-3-8B-Instruct \cite{llama3modelcard}.

For MEMIT-based baselines, we use MEMIT and an improved version of MEMIT, PMET \cite{DBLP:conf/aaai/Li0SYMY24}, AlphaEdit \cite{DBLP:conf/iclr/FangJWMSW0C25}. 
In addition to the MEMIT-based methods, we also included FT-L \cite{zhu_modifying_2020}, which was used for comparison in the ROME paper, as another baseline to verify that the same-subject issue exists only in methods with the MEMIT-based architecture.

\subsection{Results when Batch Size is 10 }
\label{subsec:same_diff_edit}

We first compared the edit success rates of standard MEMIT, PMET, AlphaEdit, \methodname, PMET-Merge, and FT-L on the two datasets.\footnote{The results for all baselines were obtained by running the code from the Easyedit framework on our datasets.}

As shown in Tab.~\ref{table:batch100_result}, our method outperforms standard MEMIT on the same-subject dataset with improved paraphrase accuracy, attributed to enhanced edit success rates. Notably, MEMIT's anomalously high specificity for same-subject edits (indicating ineffective editing and a minimal impact on the original model) is corrected by our approach, achieving specificity levels comparable to FT and distinct-subject scenarios. The results of other models are detailed in App.~\ref{sec:appendix_100batchsize_moremodel}.

Comparing the results of MEMIT, AlphaEdit and PMET to FT-L, it can be observed that the performance drop in same-subject edits is unique to MEMIT-based methods. This phenomenon is consistent with our analysis in Sec.~\ref{sec:reason_analysis}. By resolving key collisions through key-wise value merging, \methodname successfully mitigates this issue, empirically confirming that key collision is the root cause of MEMIT's limitations in same-subject cases.

The ``Merge'' method,introduced in Sec.~\ref{sec:merge-approach}, is not exclusive to MEMIT. Although PMET-Merge does not achieve the same performance as MEMIT-Merge (as illustrated in Tab.~\ref{table:batch100_result}), it can still notably mitigate performance degradation for PMET in same-subject scenarios. Our hypothesis for this difference is that PMET modifies not only the MLP but also the attention module, a dimension we did not specifically explore.

\begin{figure}[th]
    \centering
    \includegraphics[width=\linewidth]{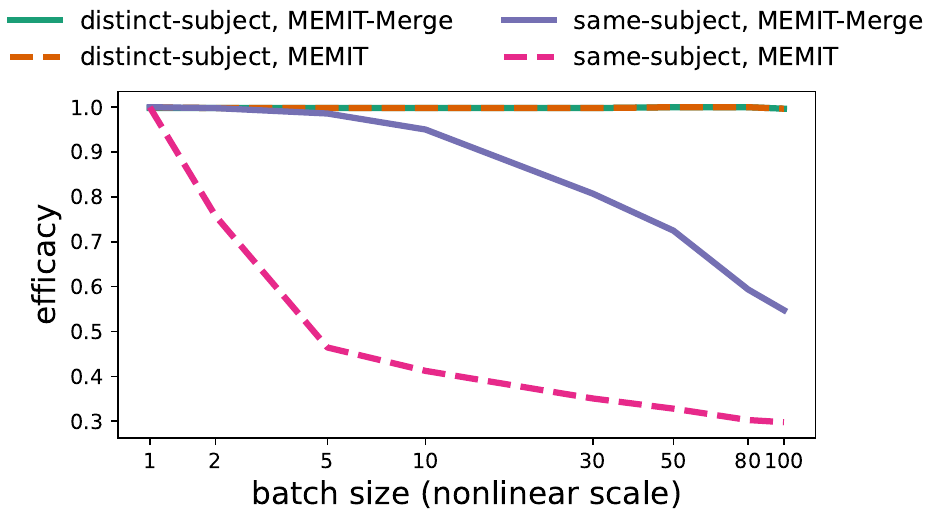}
    \caption{The results of \methodname and MEMIT methods on same-subject and distinct-subject datasets using the Qwen2.5-1.5B-Instruct model, showing the changes with varying batch sizes. \methodname is capable of significantly alleviating the decline in editing performance under the same-subject condition.}
    \label{fig:kmemit_samediff}
\end{figure}

\subsection{Results with Varying Batch Sizes}

As can be seen in Fig.~\ref{fig:kmemit_samediff}, when the subjects of the editing knowledge in the edit batch are the same, the standard edit success rate plummets at a batch size of 2, whereas \methodname is able to maintain a much higher success rate, with a significantly smaller decline compared to MEMIT. This also confirms the effectiveness of our method. The results of other models are given in App.~\ref{sec:appendix_varyingbatchsize_moremodel}

In the case of distinct subjects, the editing capability of both MEMIT and \methodname does not exhibit a significant decline even at a batch size of 100, which is consistent with our previous analysis.

It can also be observed that even when utilizing MEMIT-Merge, the outcomes for same-subject scenarios do not yet align with those achieved in distinct-subject scenarios. Although we have resolved the conflicts between key-value pairs, the capability of a value vector is not unlimited and may cause a new issue. We have enabled the value to match more relations to target objects, but performance may decline with very large batches, as a single value vector cannot handle an infinite number of associations. This is why MEMIT-Merge performs well with small batches but worsens as the batch size increases. Nonetheless, real-world editing typically involves a small number of same-subject edits (usually fewer than 10), and each subject has a limited set of properties. Given this, our method effectively addresses same-subject cases without notable performance loss.

The results of other editing methods are given in App.~\ref{sec:appendix_varyingbatchsize}.

\section{Conclusion}

This paper identifies the issue of significant performance degradation in MEMIT when a batch contains knowledge sharing the same subject during batch editing. This is fundamentally caused by parameter update conflicts arising from identical keys requiring divergent values in the same-subject scenarios. Our proposed MEMIT-Merge resolves this and significantly improves same-subject edit performance while maintaining original performance on distinct-subject cases. These findings advance mass-editing techniques for evolving LLM knowledge bases.

\section*{Limitations}
While this study provides insight into the same-subject issues within the MEMIT-based method, several limitations should be recognized.
First, all knowledge triples are restricted to person-related entities, leaving the generalization to other subject types (such as locations or organizations) untested. While our theoretical framework suggests that subject type should not fundamentally alter the conclusions, empirical validation across diverse categories remains necessary. 
Second, the experiments focus solely on lexical-level subject distinctions; potential effects of semantic similarity in embedding space were not explored. Future work could extend this investigation by incorporating larger datasets, multi-type knowledge triples, and embedding-space analyses to further validate the theoretical predictions.

Furthermore,  although we have resolved the conflicts between key-value pairs, the capability of value vector is not unlimited and may cause a new issue. The core factor, ultimately, is that the key vector at the position of the subject's last token does not contain information about the relation. Our approach is a solution that does not alter the position of the edited fact tokens. To thoroughly resolve this problem, it may be necessary to avoid using the subject's last token and instead use the last token or other positions that can incorporate the'relation' information for knowledge editing.

\section*{Acknowledgments}

This work was supported by the Natural Science Foundation of China (No. 62476134).

\bibliography{references,  anthology, custom}

\newpage

\appendix

\section{Details of Constructing Same-Subject and Distinct-Subject Data}
\label{sec:appendix_dataset_construct}

Our dataset construction is based on Wikidata. First, we retrieve all relations and properties associated with human subject entities from Wikidata. Then, we manually filter the relations, removing those that are less commonly used, such as ID and Wikidata categories. Finally, we obtain 100 relations.

Subsequently, we select a number of individuals from Wikidata and query their corresponding objects for the knowledge triples composed of these relations. Finally, we retain only one knowledge triple for each relation, thereby obtaining 100 knowledge triples, formatted as (subject, relation, object).

We then select another 100 distinct names from Wikidata and replace the subject entities in the previously obtained 100 knowledge triples with these new names, thereby creating the distinct-subject dataset. Conversely, we replace the subject entities in the 100 knowledge triples with a single, identical name to create the same-subject dataset.

Using the template ``{subject}'s {relation} is {object},'' we construct natural language sentences from these knowledge triples, which form the edit sentences in the dataset. For example, a knowledge triple in the same-subject dataset is (John Smith, doctoral advisor, Dennis W. Sciama), which is formulated into the natural sentence John Smith's doctoral advisor is Dennis W. Sciama. In the distinct-subject dataset, the corresponding knowledge triple with the same relation and object is (Paul Morand, doctoral advisor, Dennis W. Sciama), which is formulated into the natural sentence Paul Morand's doctoral advisor is Dennis W. Sciama.

Subsequently, following the dataset metrics in \citet{DBLP:conf/nips/MengBAB22}, we add two types of questions: specificity and paraphrase. For paraphrase questions, we use the same knowledge triples as the edit sentences, but with a different template format: ``The name of the {relation} of {subject} is {object}.'' For specificity, there are two types of questions. One is completely unrelated knowledge, for which we use the prompt ``The capital city of America is.'' The other type has the same relation as the edited knowledge but a different subject. For example, if the edited knowledge is (John, father, Bob), a specificity question could be (Paul, father, Eugène).

\section{Further Analysis of Cause}
\label{sec:appendix_akd_analysis}

To further investigate the relationship between the decline in editing capability and the distance between keys, we propose an evaluation metric: the Average Keys Distance Inside Batch (AKD). This metric is defined as the average Euclidean distance between the key values of all pairs of knowledge within a batch. It reflects the average distance between keys in the batch and is represented as

\begin{equation}
AKD^{(l)}=\frac{1}{\binom{|B|}{2}}\sum_{\substack{e_1\in B\\ e_2\in B}} ||k^{(l)}_{e_1}-k^{(l)}_{e_2}||_2
\end{equation}

where $l$ represents the $l$-th layer,  $B$ denotes the batch of knowledge to be edited, $k^{(l)}_{e_1}$ represents the key value computed by the MLP module in the $l$-th layer for the input knowledge $e_1$.

We compute the $AKD$ for all layers of the model at the subject's last token position.
As the degree of subject variation increases across sentences, the $AKD$ value proportionally rises. Conversely, when all sentences share identical subjects, the $AKD$ value remains constant at 0.

\begin{figure}[th]
    \centering
    \includegraphics[width=\linewidth]{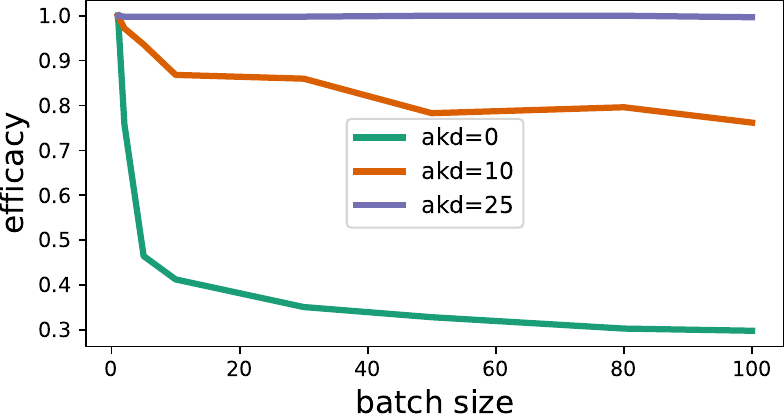}
    \caption{Datasets with different $AKD$ values and the results of edit efficacy. The lower the $AKD$ value, the more severe the decline in edit capability.}
    \label{fig:akd}
     % 一张折线图,  x 轴是 batch size,  y 轴是编辑成功率.  不同线代表不同的 $AKD$ 的数据.  （这里的 $AKD$ 值取 MEMIT 编辑的最后一层的 keys）
\end{figure}

We construct sentence batches using predefined templates, where batches sharing the same template exhibited similar $AKD$ values, while distinct templates yielded significantly different $AKD$ measurements. The specific templates and corresponding $AKD$ values are detailed in App.~\ref{sec:appendix_akd_dataset}.
For experimental validation, we select three $AKD$ groups (0, 10, 25) and conduct editing tests using Qwen2.5-1.5B-Instruct. As shown in Fig.~\ref{fig:akd}, where $AKD$ values are computed using keys from MEMIT's final editing layer, the results demonstrate an inverse relationship: lower $AKD$ values correspond to reduced editing success rates. This pattern remains consistent across other $AKD$ values, establishing a statistically significant negative correlation between $AKD$ and editing efficacy.

% \begin{figure}[th]
\section{Diverse $AKD$ Dataset}
\label{sec:appendix_akd_dataset}

\begin{table}[th]
\adjustbox{max width=\linewidth}{%
% \adjustbox{max width=\textwidth}{%
\begin{tabular}{@{}lll@{}}
\toprule
\textbf{dataset} & \textbf{formatting template} & \textbf{$AKD$} \\ \midrule
same-subject & \{subject\}'s \{relation\} is \{object\}                  & 0.0  \\
distinct-subject & \{subject\}'s \{relation\} is \{object\}                  & 25.8 \\
same-subject & The name of the \{relation\} of \{subject\} is \{object\} & 10.5 \\
distinct-subject & The name of the \{relation\} of \{subject\} is \{object\} & 26.2 \\ \bottomrule
\end{tabular}}
\caption{The average $AKD$ values obtained using different data and templates.}
\label{table:data_format_akd}
\end{table}

The construction of datasets with three distinct $AKD$ values, where the keys within each dataset have a relatively consistent distance between each other.

We utilize the knowledge triples from the same-subject and distinct-subject datasets collected in Sec.~\ref{sec:appendix_dataset_construct} to construct data using different natural language sentence templates. The two templates we employ are ``{subject}'s {relation} is {object}'' and ``The name of the {relation} of {subject} is {object}''.

Tab.~\ref{table:data_format_akd} presents the average $AKD$ values obtained using different data and templates with the Qwen2.5-1.5B-Instruct model. We selected several datasets with distinct $AKD$ values. Since these datasets have consistent internal templates, the keys of the multiple knowledge triples within them are relatively uniform and close in distance. Therefore, when performing batch editing on these datasets, they can be used to study the correlation between efficacy and $AKD$.

\begin{figure}[th]
    \centering
    \includegraphics[width=\linewidth]{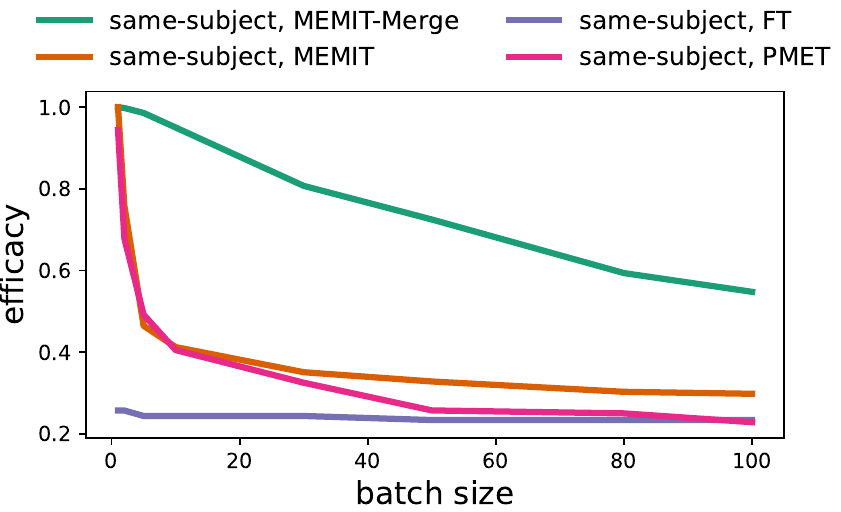}
    \caption{Editing same-subject dataset using Qwen2.5-1.5B-Instruct with four editing methods.}
    \label{fig:qwen_same_more}
\end{figure}

\section{Results of Other Models At Batch Size 10}
\label{sec:appendix_100batchsize_moremodel}

Tab.~\ref{table:more_batch100_result} shows results using Llama-3-8B-Instruct, GPT-J-6B and Qwen2.5-7B-Instruct at batch size 10.
It confirms the same-subject issue on MEMIT and the advantage of our MEMIT-Merge across different LLM model sizes and architectures.

\begin{table*}[th]
\centering
\small
\begin{tabular}{@{}llllll@{}}
\toprule
Model                                & Dataset                           & Method      & Efficacy & Paraphrase & Specificity \\ \midrule
\multirow{6}{*}{Llama-3-8B-Instruct} & \multirow{3}{*}{same-dataset}     & FT          & 0.68     & 0.63       & 0.60        \\
                                     &                                   & MEMIT       & 0.51     & 0.40       & 0.99        \\
                                     &                                   & MEMIT-Merge & 1.00     & 0.78       & 0.99        \\ \cmidrule(l){2-6} 
                                     & \multirow{3}{*}{distinct-dataset} & FT          & 0.69     & 0.65       & 0.59        \\
                                     &                                   & MEMIT       & 1.00     & 0.90       & 0.95        \\
                                     &                                   & MEMIT-Merge & 1.00     & 0.93       & 0.96        \\ \midrule
\multirow{6}{*}{Qwen2.5-7B-Instruct} & \multirow{3}{*}{same-dataset}     & FT          & 0.31     & 0.23       & 0.98        \\
                                     &                                   & MEMIT       & 0.43     & 0.37       & 1.00        \\
                                     &                                   & MEMIT-Merge & 0.99     & 0.49       & 0.99        \\ \cmidrule(l){2-6} 
                                     & \multirow{3}{*}{distinct-dataset} & FT          & 0.26     & 0.23       & 0.98        \\
                                     &                                   & MEMIT       & 1.00     & 0.82       & 0.98        \\
                                     &                                   & MEMIT-Merge & 1.00     & 0.86       & 0.97        \\ \midrule
\multirow{6}{*}{GPT-J-6B}            & \multirow{3}{*}{same-dataset}     & FT          & 0.73     & 0.57       & 0.32        \\
                                     &                                   & MEMIT       & 0.34     & 0.25       & 1.00        \\
                                     &                                   & MEMIT-Merge & 1.00     & 0.35       & 1.00        \\ \cmidrule(l){2-6} 
                                     & \multirow{3}{*}{distinct-dataset} & FT          & 0.74     & 0.69       & 0.35        \\
                                     &                                   & MEMIT       & 1.00     & 0.77       & 0.99        \\
                                     &                                   & MEMIT-Merge & 1.00     & 0.77       & 0.99        \\ \bottomrule
\end{tabular}
\caption{results of models in various size and architecture when batch size is 10.}
\label{table:more_batch100_result}
\end{table*}

\section{Results with Varying Batch Sizes of Other Models}
\label{sec:appendix_varyingbatchsize_moremodel}

\begin{figure}[th]
    \centering
    \includegraphics[width=\linewidth]{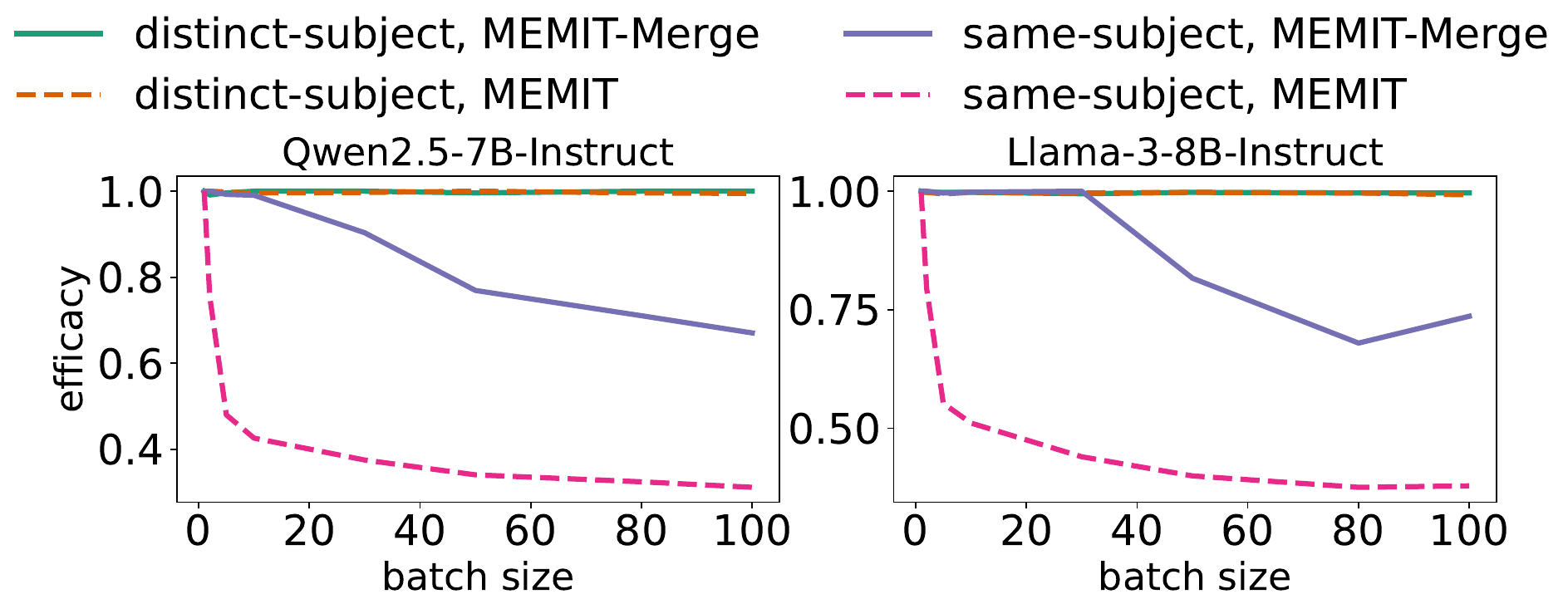}
    \caption{The results of \methodname and MEMIT methods on same-subject and distinct-subject datasets using the Qwen2.5-7B-Instruct and Llama-3-8B-Instruct.}
    \label{fig:qwen_llama_samediff}
\end{figure}

Additionally, the experimental results for Qwen2.5-1.5B-Instruct and Llama-3-8B-Instruct, two models with different architectures, as shown in Fig.~\ref{fig:qwen_llama_samediff}, demonstrate that the same phenomenon observed in the GPT-J model also exists in these models. Moreover, \methodname is equally capable of significantly mitigating the performance degradation of standard MEMIT under the same-subject condition. Therefore, it can be concluded that this phenomenon is universally present across different model architectures, and our method is applicable to various model structures.

\section{Results with Varying Batch Sizes of other methods}
\label{sec:appendix_varyingbatchsize}

Here in Fig.~\ref{fig:qwen_same_more} we demonstrate some more results about editing same subject batch with varying batch sizes.

It shows clearly that MEMIT-based methods suffer from the same subject issue, while methods like FT do not.

\section{Related Work}
\label{sec:appendix_related_work}

Knowledge editing techniques for large language models (LLMs) primarily fall into two paradigms: non-parametric approaches that preserve original parameters and parametric methods that directly modify model weights. Parametric approaches, while effective for targeted updates, often introduce uncontrolled parameter perturbations that adversely affect unrelated knowledge, a challenge addressed through various constraint mechanisms.
The parametric category features two dominant subclasses: one is ``meta-learning based methods'', such as MEND \cite{DBLP:conf/iclr/MitchellLBFM22} and MALMEN \cite{DBLP:conf/iclr/TanZF24} that train meta-networks using carefully designed datasets containing both unrelated knowledge samples and paraphrased sentences, aiming to enhance generalization while minimizing collateral damage. Another is locate-and-edit Methods, which includes techniques such as Knowledge Neuron \cite{DBLP:conf/acl/DaiDHSCW22}, identifying critical knowledge storage locations before executing precise edits. ROME \cite{DBLP:conf/nips/MengBAB22} extended this by incorporating knowledge preservation terms in its optimization objective to maintain model integrity.

Our work builds upon MEMIT \cite{DBLP:conf/iclr/MengSABB23}, a state-of-the-art locate-and-edit approach that enables batch knowledge editing through MLP layer modifications. Building on MEMIT, many recent methods have made modifications to parameter update methods during editing or to the architecture and location of the edits. PMET \cite{DBLP:conf/aaai/Li0SYMY24} incorporated the output of the attention layer in the calculation of parameter updates. AlphaEdit \cite{DBLP:conf/iclr/FangJWMSW0C25} improved upon MEMIT's parameter matrix update method by projecting the update matrix into the null space of the original knowledge to mitigate interference with unrelated knowledge. UNKE \cite{deng_unke_2024} extended structured knowledge editing to unstructured editing.

\end{document}